\documentclass[runningheads]{llncs}
\usepackage{hyperref}
\usepackage{xcolor}
\usepackage{booktabs}
\usepackage{graphicx}
\begin{document}
\title{Convolutional Neural Networks and Vision Transformers for Fashion MNIST Classification: A Literature Review}
\titlerunning{CNN and VIT for Fashion MNIST Classification : A Litterature review}
%
%
\
\author{Sonia BOUZIDI\inst{1}\orcidID{0009-0004-7876-5211} \and Ghazala HCINI\inst{1}\orcidID{0000-0003-3571-417X}\and
Imen JDEY\inst{1}\orcidID{0000-0001-7937-941X} \and
Fadoua DRIRA\inst{1}\orcidID{0000-0001-6706-4218}}
\authorrunning{Bouzidi et al.}
%
\institute{ReGIM-Lab. REsearch Groups in Intelligent Machines (LR11ES48) }
%
\maketitle              
\begin{abstract}
Our review explores the comparative analysis between Convolutional Neural Networks (CNNs) and Vision Transformers (ViTs) in the domain of image classification, with a particular focus on clothing classification within the e-commerce sector. Utilizing the Fashion MNIST dataset, we delve into the unique attributes of CNNs and ViTs. While CNNs have long been the cornerstone of image classification, ViTs introduce an innovative self-attention mechanism enabling nuanced weighting of different input data components. Historically, transformers have primarily been associated with Natural Language Processing (NLP) tasks. Through a comprehensive examination of existing literature, our aim is to unveil the distinctions between ViTs and CNNs in the context of image classification. Our analysis meticulously scrutinizes state-of-the-art methodologies employing both architectures, striving to identify the factors influencing their performance. These factors encompass dataset characteristics, image dimensions, the number of target classes, hardware infrastructure, and the specific architectures along with their respective top results.

Our key goal is to determine the most appropriate architecture between ViT and CNN for classifying images in the Fashion MNIST dataset within the e-commerce industry, while taking into account specific conditions and needs. We highlight the importance of combining these two architectures with different forms to enhance overall performance. By uniting these architectures, we can take advantage of their unique strengths, which may lead to more precise and reliable models for e-commerce applications. CNNs are skilled at recognizing local patterns, while ViTs are effective at grasping overall context, making their combination a promising strategy for boosting image classification performance.

\end{abstract}
\section{Introduction}
\begin{figure}[hbt!]
  \centering
  \includegraphics[width=12cm, height=7cm]{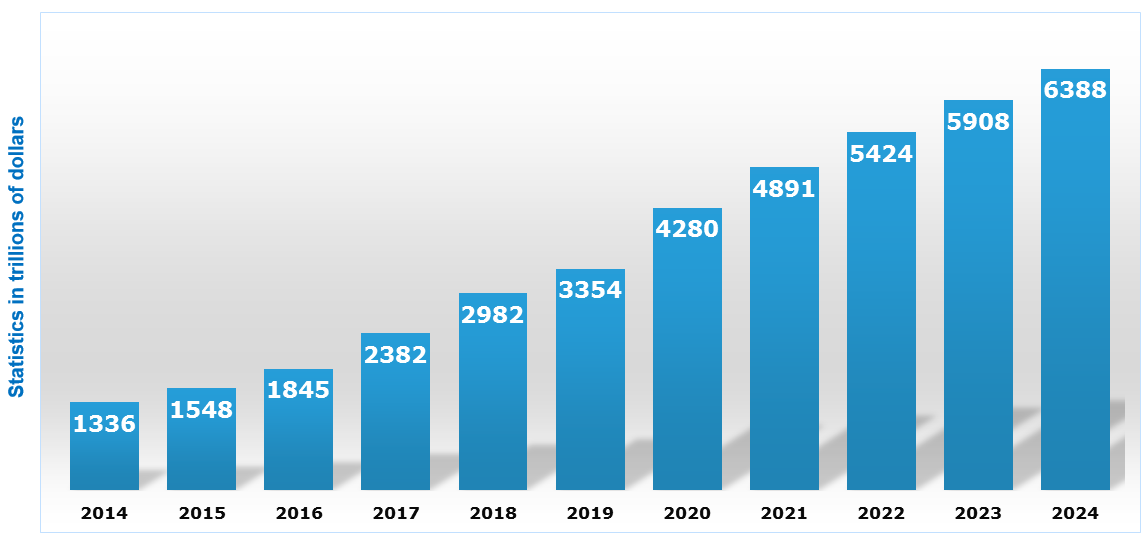}
  \caption{The evolution of online sales turnover since 2014 and predictions for 2024.}
  \label{figure1}
\end{figure}

\begin{figure}[hbt!]
  \centering
  \includegraphics[width=12cm, height=7cm]{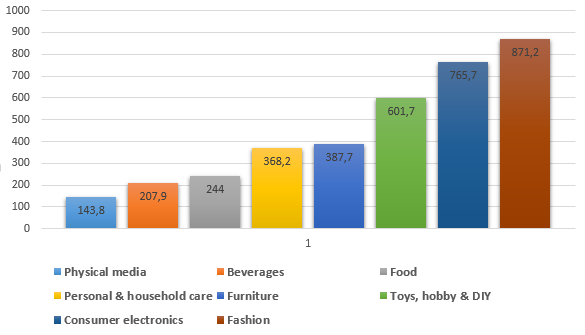}
  \caption{Estimated annual spending in each consumer goods e-commerce category in 2022.}
  \label{figure2}
\end{figure}

Presently, there is a notable surge in e-commerce sales, with the fashion sector emerging as a frontrunner, particularly in the aftermath of the COVID-19 pandemic \cite{c1}. The evolution of online sales turnover is vividly depicted in figure \ref{figure1}, showcasing a substantial rise from 1336 in 2014 to 5908 in 2023, with a promising projection of 6388 for 2024 \cite{c58}. 
Fashion is positioned as the most lucrative sector of e-commerce, holding the largest market share in 2022 among all e-commerce sectors, and now estimated at a whopping 871.2 billion in global online sales \cite{c59} as shown in figure \ref{figure2}. This financial dominance underlines the considerable impact of the fashion industry in the digital economy, highlighted by continued growth and optimistic prospects for the future.

The surge in online shopping has posed a significant hurdle for the fashion industry: the struggle to precisely distinguish garments amidst the vast array of images accessible on the internet. Despite the plethora of images offering consumers an array of options, this often leads to uncertainty regarding the true attributes of products. Consequently, this uncertainty triggers a worrying surge in product returns, impacting not only the profitability of online retail businesses but also disrupting the customer shopping experience, thereby eroding long-term brand trust and loyalty.

Employing deep learning techniques for clothing detection in online shopping endeavors to offer a pragmatic remedy to enhance consumers' online shopping encounters \cite{c2}\cite{c3}. With the flourishing e-commerce landscape in the fashion industry, this technology presents a novel avenue to cater to evolving consumer demands, demonstrating its capability to precisely discern clothing items within images.

The growing interest in Vision Transformers (ViTs) as an innovative architecture for image recognition tasks, alongside the proven track record of Convolutional Neural Networks (CNNs) in image classification, motivates this review to scrutinize and compare these two methodologies. CNNs have traditionally been lauded for their efficacy in capturing spatial hierarchies of features, serving as a cornerstone in image classification. However, the emergence of ViTs, harnessing self-attention mechanisms, has sparked considerable attention due to their capability to model long-range dependencies and adapt to diverse input sizes.

While ViTs present compelling advantages, including their capacity to handle long-range dependencies and adapt to variable input sizes, they also encounter challenges such as computational complexity, large model sizes, scalability to extensive datasets, interpretability, resilience to adversarial attacks, and generalization performance. These challenges highlight the necessity of juxtaposing ViTs with the established CNN models to gain insights into their respective strengths and weaknesses.

In this review, we outline the background in Section 2, examine related works in Section 3, provide a discussion in Section 4, and conclude in Section 5. An overview of each section is as follows:
\begin{itemize}
     \item Section 2: Background explores the performance of deep learning architectures CNN and ViT in image classification. We discuss the importance of merging these two architectures in different ways and introduce commonly used datasets and metrics crucial for classification tasks.

     \item Section 3: Related Works presents an analysis of recent studies on image classification with CNN, ViT, and their hybridization, focusing on the use of the Fashion MNIST dataset.

     \item Section 4: Discussion delves into the significance of choosing the right architecture (CNN, ViT), the value of combining these approaches, the importance of hyperparameter tuning, the specific characteristics of the fashion industry, as well as the associated limitations and challenges.

    \item Section 5: Conclusion wraps up the main points of our review and offers suggestions for future research.
\end{itemize}
\section{Background}\label{sec2}
\begin{figure}[hbt!]
  \centering
  \includegraphics[width=8cm, height=5cm]{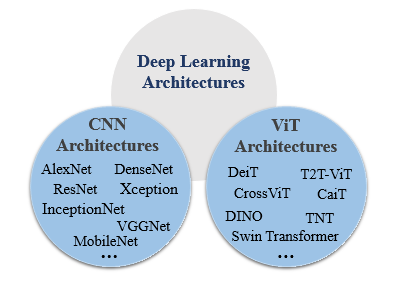}
  \caption{Different architectures of ViT and CNN.}
  \label{depf}
\end{figure}

In response to the exponential growth of data, the field of machine learning has embraced a cutting-edge approach known as "deep learning" \cite{c4}. Recognized for its exceptional performance with large datasets, deep learning has become the forefront of machine learning techniques \cite{c5}. In recent times, its application has expanded notably within computer vision tasks, particularly in the domain of fashion image classification \cite{c6}. Deep learning models, notably Convolutional Neural Networks (CNNs) and vision transformers (ViTs), have demonstrated promising results in tasks such as image classification, recognition within the fashion image \cite{c7}. Our focus remains on image classification, a crucial aspect of image processing. Similar to its application in fashion image classification, deep learning techniques are increasingly adopted in fashion image classification tasks. These techniques, including CNN, VIT and their different architectures as detailed in figure \ref{depf}, aim to automatically extract features from input images, enabling tasks like detection, segmentation, and classification without the need for manual feature engineering \cite{c8}.

Recent progress in the field of deep learning has seen an increase in research efforts focused on architectures designed for image processing. CNNs and ViTs are among the most notable models and have drawn significant interest. These architectures operate using supervised deep learning frameworks. In recent times, Vision Transformer (ViT) has gained prominence as an alternative to traditional Convolutional Neural Networks (CNNs) for image classification tasks. Research on ViTs is anticipated to grow in the future as more researchers explore this promising approach.

\begin{figure}[hbt!]
  \centering
  \includegraphics[width=12.7cm, height=6.5cm]{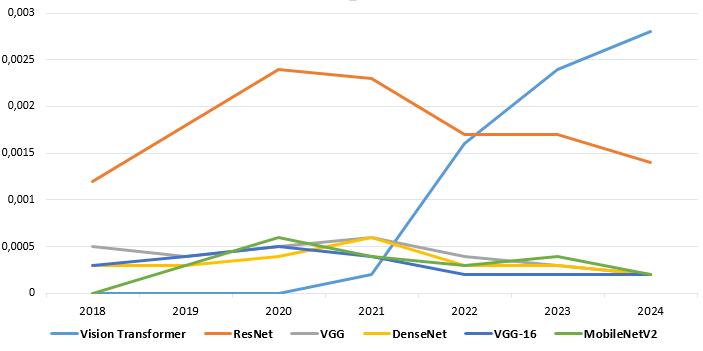}
  \caption{Number of papers using Vision Transformer and different CNN architectures in image classification tasks between 2018 and 2024.}
  \label{fig 4}
\end{figure}

To determine which model is more dominant in image classification tests, the emphasis is on the number of research publications in recent years. This method enables a comparison of different approaches based on the volume of papers published on each. The analysis follows the approach of Keele et al. \cite{c102} and includes manual searches in digital libraries such as IEEE Xplore, SpringerLink, ACM Digital Library, Wiley Online Library, and Science Direct. As shown in Figure \ref{fig 4}, transformers have gained prominence in recent years compared to various CNN methods, such as ResNet, VGG, DenseNet, VGG16, and MobileNetV2.
\newpage
Drawing from the competition between my vision transformers and convolutional neural networks (CNNs) in image classification, we aim to delineate the fundamental principles of each model.
\subsection{Convolutional Neural Networks}
\begin{figure}[hbt!]
  \centering
  \includegraphics[width=12.5cm, height=6cm]{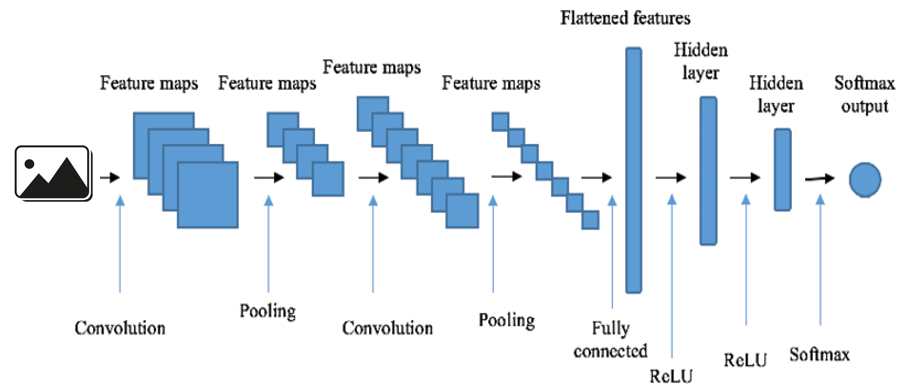}
  \caption{The standard CNN architecture.}
  \label{figure 5}
\end{figure}

    CNNs trace their origins back to the 1980s \cite{c8}, yet their widespread adoption for image classification surged following the breakthrough achievement of AlexNet in the 2012 ImageNet Large Scale Visual Recognition Challenge (ILSVRC) \cite{c9}. The underlying principle of CNNs lies in their layered architecture, with each layer playing a pivotal role in the image processing pipeline. Convolutional layers serve as the cornerstone, employing filters to discern specific patterns and structures within input images, such as edges or textures\cite{c10}. Subsequently, pooling layers typically succeed convolutional layers, facilitating the reduction of dimensionality in resultant activation maps while retaining salient features. Integrating normalization layers, like Batch Normalization, is commonplace to stabilize learning dynamics and expedite model convergence. Extracted features then traverse through fully connected layers, amalgamating spatial and structural cues to accomplish targeted objectives, such as image classification. Finally, an output layer, often manifested as a softmax layer, allocates probabilities to each conceivable class, thus enabling the classification of images based on the gleaned features \cite{c11} \cite{C103} , as shown in figure \ref{figure 5}.
The primary steps involved in CNN-based image classification encompass:
\begin{enumerate}
    \item \textbf{Convolution and Pooling:} Initially, the input image with dimensions (H×W×C) undergoes convolution with multiple filters to extract essential features. This process detects various visual patterns like edges, shapes, and textures. Following convolution, pooling operations are employed to downsize the image, making the detected features robust against minor spatial variations.
    \item \textbf{Flattening:}Subsequent to convolution and pooling, the extracted features are flattened into a one-dimensional vector. This transformation ensures that the image representation is compatible with the input requirements of fully connected layers.
    \item \textbf{Fully Connected (FC) Layer:} Flattened features are then passed through one or more fully connected layers. In these layers, each neuron is intricately connected to all neurons in the preceding layer, allowing for the extraction of intricate feature combinations.
    \item \textbf{Activation Function:} After each fully connected layer, a non-linear activation function is applied to introduce non-linearity into the model. Typically, Rectified Linear Unit (ReLU) activation is utilized, enabling the network to capture complex feature relationships.
    \item \textbf{Classification Layer:} Finally, a classification layer is incorporated, typically comprising a fully connected layer followed by a softmax function. This layer takes the extracted features and produces probabilities corresponding to each class in the classification task.
\end{enumerate}
    The most common CNN architectures are LeNet, AlexNet, VGGNet, ResNet, InceptionNet and DenseNet\cite{c87}:
    \begin{itemize}
    \item  \textbf{LeNet:} is a convolutional neural network (CNN) architecture introduced by LeCun et al. in 1998 \cite{c88}. This model played a significant role in the early development of deep learning. The most recognized version, LeNet-5, comprises seven layers: an input layer, two convolutional layers, two pooling layers, and two fully connected layers. The convolutional layers use 5x5 kernels, while the pooling layers utilize 2x2 kernels. The fully connected layers contain 84 and 10 neurons, respectively. Initially, LeNet was created for recognizing handwritten digits, but it has since been applied to other image classification tasks.
        \item \textbf{AlexNet:} Is renowned as a pioneering Convolutional Neural Network (CNN) architecture extensively employed for image recognition tasks. Unveiled in 2012, it aimed at resolving intricate large-scale image classification challenges \cite{c12}. Through its sequence of eight convolutional layers followed by three fully connected layers, AlexNet showcased the capability of deep CNNs to discern discriminative features from raw data. Its operational principle lies in the hierarchical extraction of features across multiple layers, where lower layers capture low-level features such as edges and textures, while higher layers learn complex patterns and semantic information \cite{c12}.
        \item \textbf{VGGNet:} Is distinguished as a deep Convolutional Neural Network (CNN) architecture recognized for its depth and uniformity. Introduced in 2014, VGGNet was engineered to extract intricate features from images utilizing a series of convolutional and pooling layers \cite{c13}. With its fixed-size convolutional layers (3x3) and pooling layers (2x2), VGGNet established new standards in image classification accuracy. Its operational principle involves the systematic extraction of features through multiple convolutional layers, gradually learning hierarchical representations of input data with increasing depth \cite{c14}.
        \item \textbf{ResNet:} ResNet emerges as a groundbreaking CNN architecture that introduced the concept of residual connections in 2015 \cite{c15}. Addressing the challenge of training increasingly deep networks, ResNet innovation lies in residual connections that allow information to flow directly through certain layers, mitigating the vanishing gradient problem. By enabling the training of much deeper networks (over 100 layers), ResNet revolutionized image classification performance \cite{c15}.
        \item \textbf{InceptionNet:} InceptionNet, also referred to as GoogLeNet, is a deep CNN architecture developed by Szegedy et al. in 2014 \cite{c16} \cite{c17}. The design is based on incorporating several convolutional filters of different sizes within the same layer, allowing the network to extract features at various scales. InceptionNet consists of multiple inception modules, each containing various convolutional filters of different sizes. The output from each inception module is combined and passed to the next layer. InceptionNet has been utilized for numerous image classification tasks and has achieved top performance on multiple benchmark datasets. It consists of 22 layers, including nine inception modules.
        \item \textbf{DenseNet:} DenseNet represents a CNN architecture distinguished by its dense connectivity between layers. Introduced in 2016, DenseNet offers an innovative approach where each layer is connected to all subsequent layers. This dense connectivity fosters information sharing among different layers, enhancing feature reuse and facilitating the learning of richer and more discriminative representations of input data \cite{c100}. Its operational principle involves the exploitation of dense connectivity patterns, promoting feature reuse and facilitating the flow of information throughout the network, which leads to improved learning efficiency and performance \cite{c101}.
    \end{itemize}
\begin{table}[hbt!]
\centering
\caption{Number of hyperparameters for various CNN architectures \cite{C104}.}
\begin{tabular}{|p{5cm}p{6cm}|}
\hline
Architecture & Number of Hyperparameters (M) \\
\hline
LeNet & 0.06 \\
AlexNet & 62.38 \\
VGGNet & 138.36 \\
ResNet & 25.6 \\
InceptionNet & 11.4 \\
DenseNet & 10-20\\
\hline
\end{tabular}
\label{tabhyp1}
\end{table}
 \subsection{Vision Transformers}
\begin{figure}[hbt!]
  \centering
  \includegraphics[width=12cm, height=5cm]{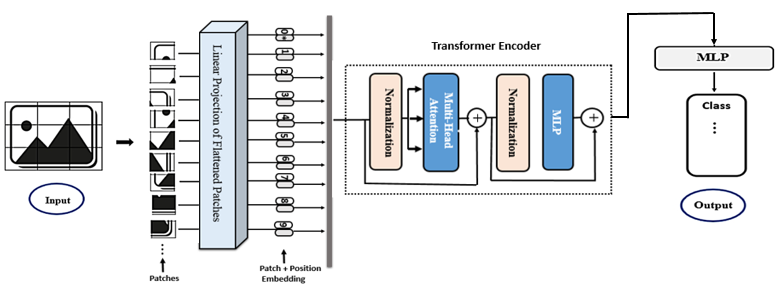}
  \caption{ViT architecture.}
  \label{figure 6}
\end{figure}

 ViT is a recent breakthrough in the area of computer vision \cite{c18}.The history of the ViT dates back to 2017, when it was initially designed for natural language processing (NLP) \cite{c36}. However, in 2020, its application expanded to computer vision tasks, marking the advent of the "vision transformer" \cite{c19}. In 2021, the ViT surpassed the Convolutional Neural Networks (CNNs) in terms of performance and efficiency, especially in image classification\cite{c19}. The ViT stands out for its ability to capture complex patterns in images thanks to its attention-based approach, thus offering an effective alternative to traditional architectures based on convolutions \cite{c20}. These advances have positioned ViT as a promising method for image classification, opening up new perspectives in the field of deep learning applied to vision \cite{c20}. The vision transformer works by dividing an image into fixed-size patches, embedding each one linearly, adding position embeddings, and then feeding the assembled vectors to a conventional Transformer encoder to create a sequence of vectors \cite{c21}. To perform classification, the traditional method involves adding a learnable "classification token" in the sequence \cite{c22} as shown in figure \ref{figure 6}.

The main Vision Transformer steps \cite{c60} \cite{c61} for image clasification are :
\begin{enumerate}
    \item \textbf{Image Patching:} The image of size (H×W×C) is divided into N patches of size (P×P×C), where H is the height, W is the width, and C is the number of channels of the image. P is the resolution of the image patch.
    \item \textbf{Linear Transformation of Patches to Vectors:} The patches are flattened so that the size of the flattened patch has a vectorized form of (1×P²*C). Each patch is transformed into a 1D vector to be used as input for the model.
    \item \textbf{Adding Position Tokens:} Position embeddings are added to the patch embeddings to retain positional information. A special classification token (CLS) is added to the positional encoding. The positional coded patch vectors are then fed as input to the Transformer Encoder.
    \item \textbf{Encoder Layer:} The transformer encoder consists of alternating layers of multi-head self-attention (MSA) and MLP (Multilayer Perceptron) blocks. A layer normalization is applied before the self-attention block and the MLP block. Residual connections are applied after each block to facilitate learning.
    \item \textbf{Classification Layer:} A classification head is implemented using an MLP with a hidden layer for feature extraction and a single linear layer for fine-tuning image classification. This layer takes as input the final representation of the image produced by the encoder and generates classification scores for each possible class.
\end{enumerate}
The most common pretrained ViT architectures a DeiT, T2T-ViT, CrossViT, CaiT, DINO, TNT and Swin Transformer \cite{c89}\cite{c94}:
\begin{itemize}
    \item \textbf{DeiT (Data-efficient image Transformers):} DeiT is a Vision Transformer that was introduced in a research paper published at ICML 2021 \cite{c90}. It is a data-efficient model that achieves comparable performance to state-of-the-art CNNs on image classification tasks \cite{c91}. DeiT uses a distillation token to transfer knowledge from a teacher model to the student model. It consists of 12 transformer layers, with a hidden size of 768 and 12 attention heads \cite{c92}.
    \item \textbf{T2T-ViT (Token-to-Token Vision Transformer):} T2T-ViT is a Vision Transformer that was introduced in a research paper published at ICLR 2021 \cite{c93}. It is a hierarchical model that uses a token-to-token structure to model images at different scales. T2T-ViT consists of 12 transformer layers, with a hidden size of 768 and 12 attention heads. It also includes a spatial reduction module to reduce the spatial dimensions of the input image.
    \item \textbf{CrossViT:} CrossViT is a Vision Transformer that was introduced in a research paper published at CVPR 2021 \cite{c95}. It is a model that uses a cross-modal embedding space to learn from both local and global features of the input image. CrossViT consists of 12 transformer layers, with a hidden size of 768 and 12 attention heads. It also includes a convolutional stem to extract local features from the input image.
    \item \textbf{CaiT (Co-scale Attention in Vision Transformers):} CaiT is a Vision Transformer that was introduced in a research paper published at ICCV 2021. It is a model that uses a co-scale attention mechanism to learn from both local and global features of the input image \cite{c96}. CaiT consists of 24 transformer layers, with a hidden size of 1024 and 16 attention heads. It also includes a hierarchical structure to model images at different scales.
    \item \textbf{DINO (Self-supervised learning of deep features for image recognition):} DINO is a self-supervised learning method for Vision Transformers that was introduced in a research paper published at ICLR 2022 \cite{c97}. It uses a teacher-student framework to learn deep features from unlabeled images. DINO consists of 12 transformer layers, with a hidden size of 768 and 12 attention heads.
    \item \textbf{TNT (Training Normalization-free Transformers):} TNT is a Vision Transformer that was introduced in a research paper published at NeurIPS 2021 \cite{C98}. It is a normalization-free model that uses a hierarchical structure to model images at different scales. TNT consists of 24 transformer layers, with a hidden size of 1024 and 16 attention heads.
    \item \textbf{Swin Transformer:} Swin Transformer is a Vision Transformer that was introduced in a research paper published at NeurIPS 2021 \cite{C99}. It is a model that uses a shifted windowing scheme to learn from the input image. Swin Transformer consists of 24 transformer layers, with a hidden size of 1024 and 16 attention heads. It also includes a hierarchical structure to model images at different scales.
\end{itemize}
\begin{table}[hbt!]
\centering
\caption{Number of hyperparameters for various ViT architectures \cite{c105}.}
\begin{tabular}{|p{5cm}p{6cm}|}
\hline
Method & Number of Hyperparameters( (M) \\ \hline
DeiT & 22.04 \\
T2T-ViT & 22 \\
CrossViT & 30.6 \\
CaiT & 27.35\\
DINO & 86 \\
TNT & 	23.8\\
Swin Transformer & 28.3 \\
\hline
\end{tabular}
\label{tabhyp2}
\end{table}
CNN and ViT each bring unique strengths and weaknesses in terms of performance and complexity. Integrating these two approaches may optimize their respective benefits and reduce their drawbacks, potentially enhancing the efficiency of image classification tasks.
\subsection{Hybrid Models}
In recent times, hybrid architectures have surfaced as a compelling strategy  to overcome the limitation of both CNNs and ViTs such as long-range dependency capture, inductive bias and rigid input size requiremen for CNN and  weak local feature extraction, sensitivity to noise and High memory cos for ViT. By combining their respective advantages, resulting in enhanced performance for image classification tasks \cite{c23}\cite{c24}. Various forms of integration, including parallel and sequential approaches, are actively being researched to optimize the synergy between these architectures. By understanding these strengths and limitations (Table\ref{tabb1} and \ref{tabb2}), researchers have devised strategies to alleviate them and engineer hybrid architectures.
\begin{table}[hbt!]
\centering
\caption{Limitations and Strengths of CNN}
\begin{tabular}{|p{5cm}p{7cm}|}
\hline
\textbf{Strengths} & \textbf{Limitations} \\
\hline
\textbf{Local Feature Extraction:} CNNs demonstrate remarkable proficiency in capturing and processing intricate local features, rendering them highly adept at tasks necessitating meticulous attention to detail, such as image recognition \cite{c62}.& \textbf{Long-Range Dependency Capture:} CNNs, limited by local receptive fields, struggle to capture extensive pixel relationships, impacting tasks like object detection and image segmentation that require understanding distant contexts \cite{c65} \cite{c66}.\\
\hline
\textbf{Inductive Biases:} This robust bias arises from the local connectivity and convolution operations, enhancing generalization performance and decreasing sensitivity to noisy data \cite{c64}. & \textbf{Inductive Bias:} CNNs demonstrate a significant inductive bias, resulting in overfitting and limited generalization. This arises as CNNs learn specific local patterns from training data, potentially hindering their ability to generalize to new, unseen data \cite{c63}.\\
\hline
& \textbf{Rigid Input Size Requirement:} CNNs often necessitate a fixed input size, reducing their adaptability to images of diverse resolutions \cite{c67}.\\
\hline
\end{tabular}
\label{tabb1}
\end{table}
\begin{table}[hbt!]
\centering
\caption{Limitations and Strengths of ViT}
\begin{tabular}{|p{5cm}p{7cm}|}
\hline
\textbf{Strengths} & \textbf{Limitations} \\
\hline
\textbf{Global Context Modeling} ViTs employ the self-attention mechanism, enabling them to capture long-range dependencies and global
contextual relationships between pixels, which is crucial for tasks like object detection and image segmentation \cite{c68}. & \textbf{Weak local feature extraction:} ViTs may struggle with tasks that require strong local feature extraction, such as image recognition tasks with fine-grained details. This is because ViTs focus on capturing global context rather than extracting fine-grained local features \cite{c70}. \\
\hline
\textbf{Scalability:} ViTs image tokenization allows adaptability to varied image sizes, offering versatility within model-defined token limits \cite{c69}. & 
\textbf{Sensitivity to noise:} ViTs can be sensitive to noisy data, as they rely on global context. Noisy data can poorly affect the self-attention
mechanism and lead to poor performance \cite{c71}.\\
\hline
 & \textbf{High memory cost:} ViTs often involve high computational costs,
especially with larger image sizes or more tokens, due to the quadratic
complexity of self-attention mechanisms \cite{c72}\\
\hline
\end{tabular}
\label{tabb2}
\end{table}
\newpage
\subsubsection{Parallel Hybridization:}
\begin{figure}[hbt!]
  \centering
  \includegraphics[width=10cm, height=4cm]{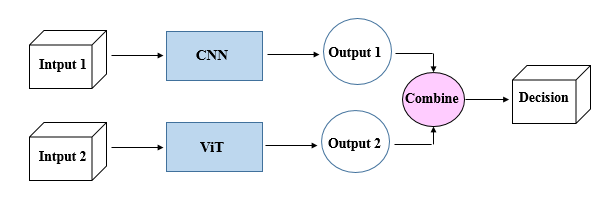}
  \caption{Parallel Hybridization.}
  \label{figure p}
\end{figure}

In parallel hybridization, both CNN and ViT networks are employed simultaneously on the same input data \cite{c23}. This strategy harnesses the spatial representation capabilities of CNNs along with ViTs' ability to capture long-range relationships within the data (Fig. \ref{figure p}). By amalgamating these distinct representations, parallel hybrid models aim to elevate both the precision and resilience of classification outcomes.
\subsubsection{Sequential Hybridization:}
\begin{figure}[hbt!]
  \centering
  \includegraphics[width=10cm, height=2.5cm]{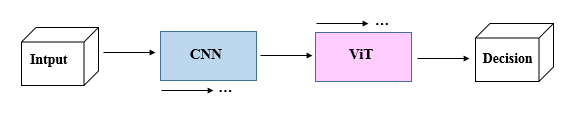}
  \caption{Sequential Hybridization.}
  \label{figure S}
\end{figure}

The sequential integration of CNNs and ViTs entails a step-by-step progression of data, where one architecture sequentially handles the input and transfers its results to the next architecture (Fig. \ref{figure S}). Usually, a CNN initiates by extracting local features, which are subsequently analyzed by the ViT to identify long-range relationships. Ensuring alignment of representations between these architectures is vital, often requiring adjustments or transformations, like resizing, reshaping, or adapting the output features from one architecture to conform to the input format demanded by the subsequent architecture during sequential processing \cite{c25}.
\subsection{Hierarchical Hybridization}
\begin{figure}[hbt!]
  \centering
  \includegraphics[width=10cm, height=2.7cm]{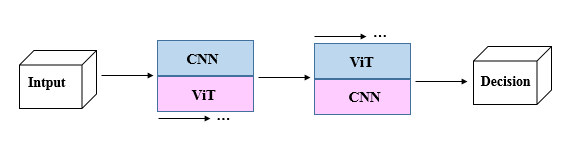}
  \caption{Hierarchical Hybridization.}
  \label{figureh}
\end{figure}

Hierarchical integration involves the fusion of Convolutional Neural Networks (CNNs) and Vision Transformers (ViTs) in a layered manner, leveraging their respective strengths at various stages of data processing(Fig.\ref{figureh}). CNNs excel in extracting local features, capturing intricate spatial details, while ViTs specialize in understanding global context and long-range dependencies \cite{c56}. Ensuring coherence and alignment between the representations of these architectures is essential in hierarchical integration. Techniques such as cross-layer connections or hierarchical fusion facilitate a smooth combination of features across different levels.

Evaluating the effectiveness of CNN, ViT, and their hybrid counterparts in image classification tasks requires careful consideration of the dataset used. The performance of these models is significantly influenced by the chosen dataset, making dataset selection a vital aspect of the evaluation process. As a result, the choice of dataset becomes a critical factor in determining the accuracy and reliability of these models in image classification applications.
\subsection{Dataset Description}
\begin{figure}[hbt!]
  \centering
  \includegraphics[width=9cm, height=6cm]{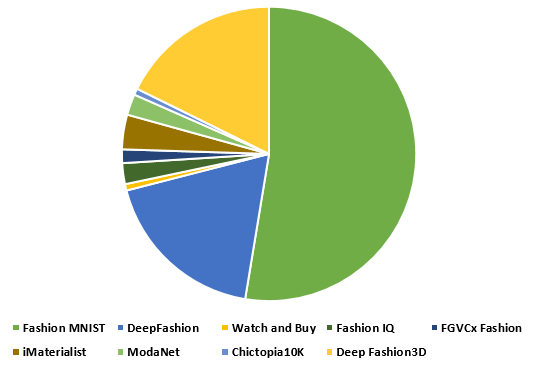}
  \caption{Number of papers using Different database of the fashion industry between 2020 and 2024 in the image classification task.}
  \label{figure3}
\end{figure}

As e-commerce continues to expand, the significance of technologies such as clothing recognition, retrieval engines, and automated product recommendation systems grows for fashion-related enterprises. However, effectively categorizing clothing presents challenges due to the diverse characteristics of garments and the complexities of classification. This difficulty extends to distinguishing between similar classes, complicating the task of multiple-class clothing classification. Consequently, the adoption of algorithms tailored to handle extensive volumes of fashion-specific data is becoming increasingly prevalent. Within the fashion industry, various datasets are utilized for image recognition as shown in table \ref{tab10}, including widely known ones such as "Fashion MNIST," "DeepFashion", "Watch and Buy" "Fashion IQ," "FGVCx Fashion", "iMaterialist", "ModaNet", "Chictopia10K", and "DeepFashion3D" \cite{c26}\cite{c27} \cite{c84} \cite{c28} \cite{c29} \cite{c30} \cite{c31} \cite{c85} \cite{c86}. 
\begin{table}[hbt!]
\centering
  \caption{Various Fashion Image Datasets. }
\begin{tabular}{|p{2.5cm}|p{3cm}|p{6.5cm}|}
 \hline
 \textbf{Dataset}& \textbf{Number of images} &\textbf{Dataset Link}\\
 \hline
  Fashion MNIST &70 000 images &\url{https://github.com/zalandoresearch/fashion-mnist} \\
  \hline
  DeepFashion & 800 000 images&\url{https://www.kaggle.com/datasets/vishalbsadanand/deepfashion-1}\\
  \hline
 Watch and Buy & 1,042,178images& \url{https://tianchi.aliyun.com/competition/entrance/531893/information}\\
    \hline
 Fashion IQ & 77 683 images&\url{https://github.com/XiaoxiaoGuo/fashion-iq}\\
  \hline
 FGVCx Fashion  &  55 000 images &\url{https://sites.google.com/view/fgvc7/home}\\
 \hline
iMaterialist  & 50 000 images&\url{https://www.kaggle.com/c/imaterialist-fashion-2019-FGVC6/overview}\\
      \hline
ModaNet  &55 000 images &\url{https://github.com/modanet/ModaNet}\\
\hline
 Chictopia10K & 17,706 images& \url{https://files.is.tue.mpg.de/classner/gp/} \\
\hline
 DeepFashion3D &2000 images & \url{https://drive.google.com/drive/folder/1JWkrjoJk7ATBhtanNm6aUOhFswRYD1WP}\\
    \hline
  \end{tabular}
   \label{tab10}
\end{table}
To find the leading dataset, we use the number of research publications over time as an important metric, enabling comparisons in image classification tasks. In line with established practices, our analysis includes manual searches across major digital libraries as outlined in section \ref{sec2}. Our results, shown in Figure \ref{figure3}, show that "Fashion-MNIST" is the most popular dataset, representing around 52.9\% of research efforts.
Fashion-MNIST is frequently employed as a benchmark to evaluate the efficacy of image classification and computer vision algorithms. Image classification serves as a fundamental aspect of computer vision \cite{c32}, extensively utilized within the fashion industry to create detailed product descriptions and streamline product tagging processes. This automated approach becomes indispensable, especially when managing vast collections, a common occurrence across various e-commerce platforms \cite{c33}.
\newpage
Fashion MNIST, curated by Zalando \cite{c73}, consists of 70,000 grayscale images representing various fashion products across 10 categories. The dataset includes a training set of 60,000 images and a test set of 10,000 images as detailed in table \ref{tab33}. Each image is a 28x28 pixel representation, associated with a label corresponding to one of the 10 clothing classes \cite{c34}.

\begin{table}[hbt!]
\centering
\caption{ Data distribution of Fashion Mnist.}
\begin{tabular}{|p{1.5cm}|p{2cm}|p{1.5cm}|p{1.5cm}|p{5cm}|}
  \hline
  Label & Description& \multicolumn{2}{c|} {Samples} & Examples \\
\cline{3-4}
 &  & Training& Test& \\
         \hline
   0  & Top & 6.000& 1.000& \includegraphics[width=1\linewidth]{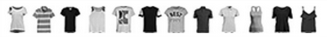}\\
   \hline
   1  & Trouser & 6.000& 1.000& \includegraphics[width=1\linewidth]{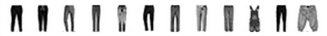}\\
    \hline
   2  & Pullover & 6.000& 1.000& \includegraphics[width=1\linewidth]{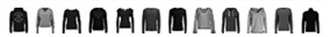}\\
    \hline
   3  & Dress & 6.000& 1.000& \includegraphics[width=1\linewidth]{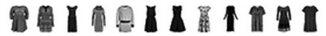}\\ 
    \hline
   4  & Coat & 6.000& 1.000& \includegraphics[width=1\linewidth]{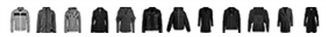}\\
       \hline
   5  & Sandal & 6.000& 1.000& \includegraphics[width=1\linewidth]{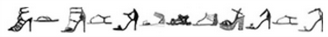}\\
       \hline
   6  & Shirt & 6.000& 1.000& \includegraphics[width=1\linewidth]{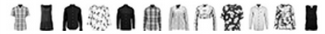}\\
          \hline
   7 & Sneaker & 6.000& 1.000& \includegraphics[width=1\linewidth]{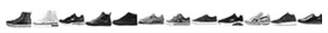}\\
\hline
   8 & Bag & 6.000& 1.000& \includegraphics[width=1\linewidth]{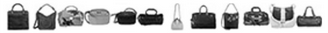}\\
\hline
   9 & Boot & 6.000& 1.000& \includegraphics[width=1\linewidth]{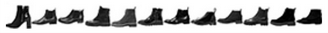}\\   
      \hline
  \end{tabular}
   \label{tab33} 
\end{table}
Fashion MNIST is evaluated using several metrics in image classification tasks. A range of measures is typically applied to assess model performance with this dataset.
\subsection{Performance Metrics}
 In the realm of image classification, multiple evaluation metrics are utilized to assess classification models notably CNN and ViT, ensuring a thorough understanding of their performance. Commonly employed metrics include accuracy, precision, recall, F1 score, and specificity, as delineated in Table \ref{tab8},  where true posi-
tive (TP), false positive (FP), false negative (FN), and true negative (TN) represent the values. This array of metrics allows for the incorporation of diverse perspectives on model performance, facilitating a more comprehensive and accurate evaluation of its classification abilities.

\begin{table}[hbt!]
\centering
\caption{Overview of Performance Evaluation Metrics.}
\begin{tabular}{p{1.5cm}p{6.5cm}p{4cm}}
\toprule
\textbf{Metrics} & \textbf{Description} & \textbf{Formula} \\
\midrule
Accuracy & Indicates the ratio of accurate predictions relative to the total number of predictions, providing insight into the model's precision in forecasting outcomes. & \begin{equation}
\frac{TP+TN}{ TP+TN+FN+FP}
\end{equation} \\
\rule{0pt}{10pt} 
Precision & Indicates the proportion of correctly predicted positive instances among all instances predicted as positive, illustrating the model's effectiveness in accurately identifying relevant outcomes. & \begin{equation}
    \frac{TP}{ TP + FP }
\end{equation} \\
\rule{0pt}{10pt} 
Recall & Denotes the fraction of samples with positive labels that were correctly identified as positive, reflecting the model's capacity to capture all relevant instances without omission. & \begin{equation}
   \frac{TP}{ TP + FN }
\end{equation} \\
\rule{0pt}{10pt} 
F1 score & Quantifies the harmonic mean of precision and recall, offering a comprehensive evaluation of a model's ability to balance accurate identification of relevant instances and minimizing false positives. & \begin{equation}
2 * ( \frac{Precision*Recall}{ Precision+Recall })
\end{equation} \\
\rule{0pt}{10pt} 
specificity & Quantifies the fraction of true negative instances correctly identified as negative, providing an indication of the model's capability to accurately discern negative outcomes and minimize false positives & \begin{equation}
  \frac{TN}{ TN + FP }
\end{equation} \\
\bottomrule
\end{tabular}
\label{tab8}
\end{table}
\newpage
After outlining the background and foundational concepts, we will now review recent developments in image classification research. The next section focuses on key works that utilize CNN, ViT, and hybrid methods with the Fashion MNIST dataset.
\section{Related Works}
\begin{figure}[hbt!]
  \centering
  \includegraphics[width=6cm, height=5cm]{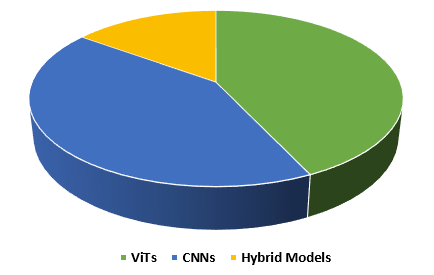}
  \caption{Number of papers using  CNNs, ViTs and hybridizing CNN and ViT for Fashion MNIST Classification.}
  \label{fig 7}
\end{figure}

The following search terms were used to find papers on artificial intelligence and Fashion MNIST dataset, we applied the following two research methods :\\
\textbf{C1:} "Deep learning", "Vision Transformer (ViT)", and "Fashion MNIST classification”.\\
\textbf{C2:} "Deep Learning" and "Convolutional Neural Networks (CNN)” and "Fashion MNIST Classification".\\
\textbf{C3:} "Deep Learning" and "Hybridizing Vision Transformer(ViT) and Convolutional Neural Networks(CNN)” and "Fashion MNIST Classification"(Figure \ref{fig 7}).
\newpage
\subsection{CNNs for Fashion MNIST Classification}
Leithardt et al. \cite{c35} developed garment recognition algorithms to meet the rising demands of the online fashion market. Their study utilized Convolutional Neural Network (CNN) models to classify fashion items using the Fashion-MNIST dataset. They observed a significant improvement in accuracy, from 89.7\% to 99.1\%, achieved with a novel CNN model termed cnn-dropout-3. This research highlights the potential of CNN models in accurately classifying fashion items, thereby enhancing sales strategies in the online fashion market.

Khanday et al. \cite{c36} investigated the impact of filter size hyperparameters on Convolutional Neural Networks (CNNs) for image classification, specifically focusing on the FashionMNIST dataset. They explored three filter sizes (3×3, 5×5, and 7×7) while maintaining a constant number of filters (32). The study found that smaller filter sizes led to higher classification accuracy, with the 3×3 filter achieving an accuracy of 93.68\% on the FashionMNIST dataset. However, it was noted that smaller filter sizes resulted in longer training times due to increased computational costs. These findings emphasize the significance of filter size selection in optimizing CNN performance for image classification tasks, particularly in datasets like Fashion MNIST.

Nocentini et al. \cite{c37}explored the use of neural network models to improve apparel image classification accuracy on the Fashion-MNIST dataset, particularly in the context of service robotics for elderly care. Their study focused on addressing the challenge of clothing manipulation, crucial for assisting elderly individuals in household settings. By proposing four different neural network models and testing them on various datasets, including Fashion-MNIST and customized datasets, Nocentini et al. achieved significant advancements. Notably, their model, the Multiple Convolutional Neural Network with 15 convolutional layers (MCNN15), outperformed previous literature with a classification accuracy of 94.04\% on the Fashion-MNIST dataset.This research underscores the importance of advanced neural network models in effectively addressing practical challenges, such as object-based manipulation tasks in household environments, thus contributing to the development of service robotics for eldercare.

Erkoç et al.\cite{c38}introduced a unique training algorithm for convolutional layers in CNNs, operating without supervision or backpropagation. This method facilitates feature extraction from unlabeled data, automatically determining the optimal number of filters post-filter extraction and eliminating the need for weight initialization mechanisms. The approach simplifies CNN training, reducing reliance on labeled data and hyperparameter tuning. Impressively, on datasets like MNIST, EMNIST-Digits, Kuzushiji-MNIST, and Fashion-MNIST, the method achieved high test performances of 99.19\%, 99.39\%, 95.03\%, and 90.11\%, respectively, without using backpropagation or preprocessed data.

Swain et al. \cite{c39} introduced a computer vision solution employing Convolutional Neural Networks (CNN) and the LeNet model to assist visually impaired individuals in recognizing clothing items. They utilized the Fashion-MNIST dataset for training and validation. The CNN model attained an accuracy of 93.7\%, while the LeNet model demonstrated superior performance with an accuracy of 98.4\%. These results underscore the effectiveness of deep learning techniques in facilitating clothing classification tasks for individuals with visual impairments.

Yu et al. \cite{c40} introduced a novel method for clothing classification utilizing a convolutional neural network (CNN) architecture "FFENet( frequency-spatial feature enhancement network for clothing classification). Their approach incorporates a Discrete Cosine Transform (DCT) frequency domain enhancement module to extract spatial information from various frequency bands of the image. These spatial feature maps are then merged to create a comprehensive feature map, which undergoes further processing via a clothing feature extraction module. The method concludes with a sequence of operations, including a 1 × 1 convolution, global average pooling, and two fully connected layers, culminating in the final clothing classification outcomes. Notably, the method achieved an accuracy of 94.62\% on the Fashion MNIST dataset and demonstrated promising performance on the Clothing-8 dataset.

Wan et al. \cite{c41} introduce LMFRNet, a lightweight convolutional neural network (CNN) model tailored for resource-constrained environments. The innovation of LMFRNet lies in its multi-feature block design, which effectively reduces model complexity and computational load while maintaining high performance. Remarkably, achieving an accuracy of 94.6\% on the CIFAR-10 dataset, LMFRNet demonstrates remarkable efficiency in image analysis tasks. Furthermore, the authors validate the model's performance across various datasets, including CIFAR-100, MNIST, and Fashion-MNIST, showcasing its robustness and adaptability. Particularly noteworthy is LMFRNet's accuracy of 94.11\% on the Fashion-MNIST dataset, highlighting its suitability for real-world applications with limited computational resources.

Sun et al. \cite{c42} advanced MADPL-net, an innovative end-to-end model integrating CNN learning, deep encoder learning, and attention dictionary pair learning (ADicL) in a unified framework. MADPL-net addresses limitations of previous methods by jointly updating dictionary learning and classification, enhancing classification performance. The incorporation of ADicL enables selection of image-attentive atoms, improving classification capability. Empirical results demonstrate MADPL-net's superiority over existing methods, achieving an impressive 91.24\% accuracy.

The approach devised by Shin et al. \cite{c43}for the classification of clothing images utilizes an enhanced CNN architecture, integrating refined convolutional and pooling layers activated by ReLU, along with prudent management of dropout layers to mitigate overfitting risks. They introduced a novel dynamic learning rate approximation method, facilitating automatic adjustment during training, consequently reducing training duration and enhancing convergence. Performance evaluation conducted on the Fashion-MNIST dataset yielded a notable classification accuracy of 93\%, marking a 4.6\% advancement compared to the baseline CNN model. This innovative methodology offers a promising avenue for clothing image recognition tasks

In their research, METLEK et al.\cite{c44} introduce CNNTuner, a novel model designed for clothing image classification. They utilize the Fashion MNIST dataset as the basis for their evaluation. The CNNTuner architecture consists of 15 layers, with hyperparameter optimization achieved through the integration of the Keras Tuner tool. Important parameters like the number of filters in convolutional layers, window sizes, and activation functions are automatically selected to ensure seamless compatibility across different layers. Through experimentation, the authors achieve remarkable performance metrics, including F1 scores, recall, precision, specificity, and accuracy, all exceeding 0.99 or 99.18\%. This automated approach represents a significant advancement in clothing image classification, with potential applications in tasks such as clothing search and product recommendation.

The table \ref{tab6} summarizes the works mentioned above
 \begin{table}[hbt!]
\centering
  \caption{Existing CNNs Approaches in Literature}
\begin{tabular}{|p{1.5cm}|p{2cm}|p{4.5cm}|p{3.5cm}|}
 \hline
 \textbf{Years}&\textbf{References}& \textbf{Methods}&\textbf{Performances}\\
  \hline
2021 &\cite{c35} & CNN-dropout-3 & Accuracy: 99.1\%
\\
   \hline
 2021&\cite{c36} & CNN & Accuracy: 93.68\% \\
   \hline
2022 &\cite{c37} & MCNN15 & Accuracy: 94.04\% \\
   \hline
2023 &\cite{c38} & Unsupervised Filter Learning &Accuracy: 90.11\%\\
   \hline
2023 &\cite{c39} & \begin{itemize}
     \item CNN
     \item LeNet
 \end{itemize} &\begin{itemize}
     \item Accuracy: 93.7\%
     \item Accuracy: 98.4\%
 \end{itemize}  \\
 \hline
2023 &\cite{c40} & FFENet & Accuracy: 94.62\%
 \\ 
  \hline
2023 &\cite{c41} & LMFRNet  & Accuracy: 94.11\% \\ 
  \hline
 2023 &\cite{c42} & MADPL-net & Accuracy: 91.24\% \\
    \hline
2023& \cite{c43} & CNN & Accuracy: 93\%  \\
     \hline
2023& \cite{c44} & CNNTuner &\begin{itemize}
     \item Accuracy: 99.18\% 
     \item Precision: 99\%
     \item F1 score: 99\%
     \item Recall: 99\%
     \item Specificity: 99\%
     \item F2 score: 99\%
 \end{itemize}  \\
    \hline
  \end{tabular}
   \label{tab6}
\end{table}
\subsection{ViTs for Fashion MNIST Classification}

Mukherjee et al. \cite{c45} developed a groundbreaking deep learning framework, OCFormer, based on Vision Transformers (ViT), specifically designed for one-class classification tasks. Unlike previous methods that employed multiple loss functions, OCFormer simplifies the process by utilizing zero-centered Gaussian noise as a surrogate pseudo-negative class in the latent space representation, optimizing the training process with a specialized loss function. Through extensive experimentation on various datasets such as CIFAR-10, CIFAR-100, Fashion-MNIST, and CelebA eyeglasses datasets, the effectiveness of OCFormer was demonstrated. It outperformed conventional CNN-based one-class classification techniques, achieving an impressive accuracy of 92.71\%.

Chhabra et al. \cite{c46}  offered PatchRot, a self-supervised technique tailored for vision transformers. PatchRot involves rotating images and image patches and training the network to predict these rotation angles, enabling the network to extract both global and local features effectively. Through extensive experimentation on various datasets, PatchRot training was found to learn rich features surpassing those achieved by supervised learning and benchmarking. This approach yielded an accuracy of 92.6\%.

Chhabra et al. \cite{c47} proposed a regularization technique called PatchSwap, where patches are exchanged between two images, creating a fresh input for transformer regularization. Through comprehensive experimentation, it was found that PatchSwap outperforms existing advanced methods. Notably, its straightforward implementation facilitates a smooth transition to semi-supervised learning environments with minimal complexity. Impressively, utilizing PatchSwap achieved an accuracy of 92.6\%, highlighting its effectiveness in improving model performance.

Bacochina et al. \cite{c48} introduced a new attention mechanism called "Element-Wise Attention Layers," which aims to optimize Dot-Product Attention by transitioning from matrix to element-wise operations. This approach was evaluated on Fashion MNIST and CIFAR10 datasets using models with VGG-like architectures. Results demonstrate a significant improvement in parameter efficiency, with the proposed mechanism achieving 92\% accuracy on Fashion MNIST while reducing parameter count by 97\%. For CIFAR10, accuracy remains competitive at 60\% of the VGG-like counterpart while utilizing 50\% fewer parameters. This highlights the effectiveness of the element-wise attention approach in enhancing model performance and efficiency.

Abd Alaziz et al. \cite{c49} conducted a study with the aim of refining fashion image classification techniques. Their research introduced a novel methodology employing Vision Transformer (ViT) architecture, which integrates transformer blocks comprising self-attention and multilayer perceptron layers. The authors meticulously fine-tuned the hyperparameters of the ViT model to maximize its efficacy. Comparative assessments were conducted with other models, encompassing various CNN architectures such as VGG16, DenseNet-121, Mobilenet, and ResNet50, across two distinct fashion image datasets. Evaluation metrics, including accuracy, precision, recall, F1-score, and AUC, were employed to gauge model performance. The findings underscored ViT's superiority over alternative models, illustrating its effectiveness in fashion image classification tasks. For instance, on the Fashion-MNIST dataset, ViT achieved notable results with an accuracy of 95.25\%, precision of 95.20\%, recall of 95.25\%, and an F1-score of 95.20\%. This extensive evaluation highlights ViT's significant potential in advancing fashion image classification methodologies.

Rodriguez et al. \cite{c50} proposed a method to enhance privacy in image classification by employing two different image transformation techniques: one utilizing convolutional autoencoder (CAE) latent representation and the other utilizing vision transformer (ViT) embeddings. These approaches are designed to develop classification models while concealing sensitive image information and improving resistance against reconstruction attacks. We assessed their performance across multiple datasets including Fashion MNIST, CIFAR-10, and Chest X-ray. The CAE-based method achieved a classification accuracy of 91.43\% on Fashion MNIST, while the ViT-based approach attained 87.32\%. These findings underscore the effectiveness of both strategies in maintaining model utility while bolstering privacy measures.

Shah et al. \cite{c51} introduced a novel method called SWEKP-based ViT (Stochastic Weighted Composition of Contrastive Embeddings \& Divergent Knowledge Dispersion for Heterogeneous Patch Encoding), aimed at enhancing Vision Transformers (ViT) for image recognition tasks. This approach modifies the patch encoding module to create heterogeneous embeddings, incorporating traditional linear projection and positional embeddings along with three additional embeddings: Spatial Gated, Fourier Token Mixing, and Multi-layer perceptron Mixture embedding. Additionally, a Divergent Knowledge Dispersion (DKD) mechanism is proposed to efficiently propagate latent information across the transformer network. The performance of the model was evaluated on four benchmark datasets: MNIST, Fashion-MNIST, CIFAR-10, and CIFAR-100. Notably, on the Fashion-MNIST dataset, the proposed method achieved an accuracy of 93.57\%, representing an improvement over conventional ViT models.

Xu et al. \cite{c52} introduce a groundbreaking method named Optronic Vision Transformer (OPViT), which integrates spatial light modulators (SLMs) and lens groups to facilitate matrix multiplication, heralding the introduction of Transformer into optical neural networks. This innovative approach maximizes the advantages of photonic computing, leading to a significant reduction in computational costs. OPViT exhibits reconfigurability, scalability, and low space complexity. Remarkably, with just one layer of Transformer and two or three layers of optical convolution, OPViT achieves remarkable classification results. The authors evaluate this approach on both the MNIST and Fashion-MNIST datasets, achieving a test accuracy of 88.93\% on the Fashion-MNIST dataset, surpassing existing optical convolutional neural networks and rivalling previous optical architectures connected to electronic networks.

The Table \ref{tab7}, provides a concise overview of the previously discussed works.
 \begin{table}[hbt!]
\centering
  \caption{Existing CNNs Approaches in Literature}
\begin{tabular}{|p{1.5cm}|p{2cm}|p{4.5cm}|p{3.5cm}|}
 \hline
 \textbf{Years}&\textbf{References}& \textbf{Methods}&\textbf{Performances}\\
 \hline
 2022&\cite{c45} & OCFormer & Accuracy: 92.71\%  \\
   \hline
2022 &\cite{c46} & PatchRot &Accuracy: 92.6\% \\
    \hline
 2022&\cite{c47}& PatchSwap & Accuracy: 92.6\% \\
   \hline
 2023&\cite{c48} & Element-Wise Attention Layers &  Accuracy: 92.87\%  \\
 \hline
 2023&\cite{c49} &ViT & \begin{itemize}
     \item Accuracy: 95.25\% 
     \item Precision: 95.20\%
     \item Recall: 95.25\%
     \item F1-score: 95.20\%
     \end{itemize} \\
     \hline
    2023&\cite{c50} &  ViT&  Accuracy: 87.32\% \\
     \hline
   2023 & \cite{c51} & SWEKP-ViT& Accuracy: 93.57\% \\
          \hline
     2023&\cite{c52} & OPViT&  Accuracy: 88.93\%\\
    \hline
  \end{tabular}
   \label{tab7}
\end{table}
\subsection{Hybrid Models for Fashion MNIST Classification}
Shao et al. \cite{c53} have introduced a groundbreaking visual transformer architecture called TSD, which blends convolutional neural networks (CNNs) and transformers using novel blocks such as CPSA and LFFN in an parallel Integration. This innovative design captures both local and long-range image details effectively. Experiments demonstrated that training the TSD model from scratch on small datasets resulted in exceptional performance while keeping the model lightweight and efficient. The TSD model outperformed existing CNN-based and transformer-based models, reaching a top-1 accuracy of 96.56\% on the Fashion MNIST dataset with fewer computations and parameters. The researchers found a way to eliminate the need for positional token encoding without sacrificing model performance, simplifying the design and allowing for greater flexibility in adjusting input image resolutions. This adaptability is crucial for various vision-based applications. Overall, the TSD model achieved an outstanding accuracy of 96.56\% on the Fashion MNIST dataset.

 Cools et al.\cite{c54} introduced a pioneering hybrid architecture known as CAReNet (Convolutional Attention Residual Network), which integrates convolutional neural networks (CNN) with attention mechanisms in an Hierarchical integration for image classification tasks. This architecture was developed from scratch to mitigate architectural disparities often encountered with pre-trained models. CAReNet comprises an initial convolutional block for fundamental feature extraction, followed by parallel blocks incorporating bottleneck structures and residual units to enhance feature extraction and information flow. These features are merged through element-wise addition, followed by max pooling for spatial hierarchy refinement. This modular approach is iterated thrice with varying channel dimensions to capture multi-scale features. The architecture also includes a CareNet block featuring bottleneck structures and attention mechanisms operating in grid and window patterns, followed by feature averaging for consolidation and consistent propagation through an additional residual unit. CAReNet demonstrates competitive performance across benchmark datasets, achieving notable accuracy, including an impressive 95\% top1 validation accuracy on the Fashion-MNIST dataset, showcasing its efficacy in diverse image classification tasks.

Meng et al. \cite{c55} have presented an innovative method called MixMobileNet that blends convolutional neural networks (CNNs) with vision transformers (ViTs) in an architectural hybrid to enhance both local and global image feature extraction. This technique resulted in a high accuracy of 95.37\% on the Fashion MNIST dataset. MixMobileNet introduces MMb (MixMobile) blocks, which encompass an LFAE (Local-Feature Aggregation Encoder) for extracting local features and a GFAE (Global-Feature Aggregation Encoder) for capturing global features. The GFAE encoder optimizes computation by reducing the channel dimensions and applying feature attention calculations per channel. Meanwhile, the LFAE encoder uses adaptive PConv (Partial-Conv) convolutions to capture local features at different scales within an image. MixMobileNet has demonstrated versatility and effectiveness in image classification across multiple datasets.

Xu et al. \cite{c57} have introduced a novel method called HSViT (Horizontally Scalable Vision Transformer) to address key obstacles in computer vision. By merging the strengths of convolutional neural networks (CNNs) with Vision Transformer architectures (ViT) in an Hierarchical Integration, HSViT resolves the lack of pre-training on large datasets and limitations of mobile computing resources. This approach uses image-level embeddings to allow ViT to leverage the inductive bias of CNNs. The horizontally scalable architecture minimizes layers and parameters and enables collaborative training across multiple nodes. Experiments, particularly with Fashion-MNIST, show HSViT achieving up to 10\% higher top-1 accuracy compared to other methods, making it a promising solution for efficient learning and inference across different platforms.

The table \ref{tab9} summarizes the works mentioned above
 \begin{table}[hbt!]
\centering
  \caption{Existing Hybrid ViT with CNN Approaches in Literature.}
\begin{tabular}{|p{1cm}|p{1.8cm}|p{2cm}|p{3.5cm}|p{3.5cm}|}
 \hline
 \textbf{Years} &\textbf{References}& \textbf{Methods} & \textbf{Types of hybridization}&\textbf{Performances}\\
 \hline
 2022&\cite{c53} & TSD& Parallel Integration & Accuracy: 96.56\%
   \\
  \hline
 2023&\cite{c54} & CAReNet & Hierarchical Integration &
      Top1 validation accuracy: 95\%
  \\
    \hline
   2024 & \cite{c55} & MixMobileNet &Sequential Integration &
      accuracy: 95.37\%  \\
      \hline 
     2024 & \cite{c57} &HSViT & Hierarchical Integration & Top-1 Accuracy: 95.92\%\\
    \hline
  \end{tabular}
   \label{tab9}
\end{table}

\section{Discussion}
In this section, we delve into various essential aspects, including the significance of architectural choices by comparing the advantages and drawbacks of CNN and ViT models. We consider fusion methods that leverage the strengths of both architectures and the role of hyperparameter tuning for optimizing both CNN and ViT models performance. Additionally, we examine the practical uses of these models in the fashion industry, such as clothing classification, recommendation systems, virtual try-on technologies, and trend analysis. Lastly, we discuss existing limitations and challenges, such as data scarcity, domain adaptation, and model interpretability, offering insights into possible future research directions and areas for improvement.
\subsection{Importance of Architecture Choice (CNN, ViT)}
When considering image classification with the Fashion MNIST dataset, the selection of architecture, CNNs or ViTs is pivotal to achieving optimal model performance. Both architectures have shown their strengths and weaknesses in various approaches.

CNN-based methods, such as CNN-dropout-3 and CNNTuner, have achieved impressive accuracy rates of 99.1\% and 99.18\% respectively. Their success is largely due to their ability to effectively capture localized features in images and adapt well to image classification tasks. However, other CNN strategies, like the standard CNN methods (93.68\% and 94.04\% accuracy), have shown slightly lower outcomes.

ViTs also demonstrate robust results in image classification. Strategies such as SWEKP-ViT and ViT have yielded accuracies of 93.57\% and 95.25\% respectively. ViTs leverage attention mechanisms to identify global patterns in images, which may lead to stronger generalization and precision.

That said, the performance of ViT methods can vary across different models, as evidenced by the range of accuracies observed (87.32\% to 95.25\%). These variations highlight the importance of choosing a method tailored to the specific task and dataset.

Overall, both CNNs and ViTs have distinct advantages in classifying images in the Fashion MNIST dataset. CNNs are a reliable option due to their strong track record and adaptability, while ViTs bring fresh perspectives with their use of attention mechanisms. There is potential in combining CNNs and ViTs to create hybrid models that can capitalize on the strengths of both architectures using CNNs for effective local feature extraction and ViTs for broader contextual understanding.

\subsection{Fusion Approaches}
Hybrid models that integrate CNNs and ViTs have demonstrated promising potential in image classification tasks. These approaches merge the local feature extraction capabilities of CNNs with the broader context understanding offered by ViTs attention mechanisms. For instance as shown in table \ref{tab9}, TSD achieved an accuracy of 96.56\%, showcasing the benefits of combining the two architectures. Similarly, CAReNet and MixMobileNet reached accuracies of 95\% and 95.37\% respectively, and HSViT achieved a Top-1 accuracy of 95.92\% highlighting the reliability of these models.

 Fusing CNNs and ViTs can create models that are more robust and adaptable for various image classification tasks. As research in this area progresses, these hybrid approaches could become crucial for enhancing image classification performance, offering new and promising directions for visual AI. However, it's important to acknowledge that while this fusion is indeed effective, it brings along its own set of challenges. One such challenge is the increased complexity and computational costs associated with integrating two distinct architectures. CNNs and ViTs, both inherently complex models, when combined, substantially escalate the computational requirements. This heightened complexity may pose significant hurdles, particularly in resource-constrained settings, necessitating more powerful hardware and potentially elongating the training process. Furthermore, the amalgamation of these architectures could potentially obscure model interpretability and lengthen development cycles, as ensuring seamless integration and optimization may demand additional time and effort.

\subsection{Hyperparameters Selection}
In deep learning, parameters and hyperparameters  play a distinct role in shaping and training models \cite{c80}. Parameters are the variables that are learned and adjusted by the model during training, such as weights and biases in neural networks \cite{c81}. These values change as the model is trained to minimize the loss function value and improve its performance on the task at hand.

Hyperparameters, on the other hand, are the settings that define the architecture and training process of the model \cite{c82} \cite{c83}. They include aspects such as learning rate, batch size, the number of layers, and architecture-specific options like patch size in ViTs or filter size in CNNs. Unlike parameters, hyperparameters are not learned during training; instead, they are preset before training begins. They play a crucial role in the overall behavior and efficiency of the model, impacting its learning speed, accuracy, and ability to generalize to new data. Choosing appropriate hyperparameters is key to achieving the best possible performance from a model.
\subsubsection{CNN Hyperparameters:}
CNNs rely heavily on hyperparameters to shape their architecture and tune performance in image classification tasks. These hyperparameters influence various aspects of the model, affecting the model's ability to learn and generalize effectively. The following are some instances of how changing hyperparameters might improve model performance:
\begin{itemize}
\item Number of layers: The intricacy and capacity of a neural network can be influenced by the count of hidden layers within the model. More hidden layers typically allow the network to learn more complex patterns and relationships in the data.
\item Activation functions: are crucial for transforming the input each neuron receives, enabling the model to learn better representations. For example, Rectified Linear Unit (ReLU) and the softmax function are frequently employed in deep learning due to their effectiveness and their ability to address the vanishing gradient issue.
\item Learning rate: determines the step size at which the model updates its weights during training. It is important to choose an appropriate learning rate; a high rate can cause the model to converge too quickly or overshoot the optimal solution, while a low rate may lead to slow convergence. 
\item Optimizers: such as Stochastic Gradient Descent (SGD), Adam, Adagrad, and RMSprop adjust the model's weights and biases based on the gradient of the loss function. Each optimizer has its own set of hyperparameters that can be fine-tuned to improve model performance.
\item Number of Epochs: refers to the number of times the model goes through the entire training dataset. Increasing the number of epochs can enhance the model's performance, but excessive training can result in overfitting, where the model performs well on the training data but struggles with new, unseen data.
\item Dropout Rate: helps prevent overfitting by randomly deactivating connections during training, promoting better generalization.
\item Kernel Size: affects the receptive field of convolutional layers, impacting the model's ability to capture features from the input data. Larger kernels can capture broader patterns, while smaller kernels focus on finer details. The choice of kernel size should align with the dataset's characteristics and the features of interest.
\item Batch Size: Refers to the number of data samples the model processes at once during training. It affects the speed of learning and the amount of memory needed. Choosing the right batch size helps balance training efficiency and model stability.
   \end{itemize} 
Fine-tuning hyperparameters is crucial for optimizing the performance and efficiency of CNN architectures. Selecting the best hyperparameter values can significantly impact how effectively a model learns from data, its convergence speed, and its ability to generalize to new data. By adjusting these settings, one can customize a model's behavior to suit a specific task or dataset, leading to improved accuracy and robustness. Table \ref{tab11} provides a detailed look at the hyperparameters used in various CNN architectures with the Fashion MNIST dataset. This analysis enables researchers and practitioners to make thoughtful decisions when building models, helping to maximize performance and achieve optimal results.
 \begin{table}[hbt!]
\centering
  \caption{Hyperparameters of CNN architectures in various studies}
\begin{tabular}{|p{3cm}|p{3cm}|p{5cm}|}
 \hline
 \textbf{Hyperparameters }& \textbf{Values}&\textbf{References}\\
 \hline
 Batch size &123, 8, 128 & \cite{c35}, \cite{c40}, \cite{c43}\\
 \hline
 Number of layers & 2, 13, 5, 3, 15 & \cite{c35} \cite{c39}, \cite{c36}, \cite{c37}\cite{c38}, \cite{c41}, \cite{c44}\\
  \hline
 Activation functions  & SoftMax, ReLU and SoftMax, ReLU & \cite{c35}, \cite{c36}, \cite{c37} \cite{c38}\cite{c39} \cite{c43} \cite{c44}, \cite{c41} \\
    \hline
Learning rate  & 0.001, 0.01, 0.1 &\cite{c36} \cite{c37}\cite{c39}, \cite{c40}, \cite{c41} \cite{c43}  \\
    \hline
Optimizers  &Adadelta, Adam,  SGD, Adam and SGD & \cite{c35} \cite{c38}, \cite{c36} \cite{c37} \cite{c39} \cite{c44}, \cite{c40}, \cite{c41} \\
    \hline
 Number of Epochs & 12, 50, 20, 100 & \cite{c35}, \cite{c38}, \cite{c44}, \cite{c37} \\
    \hline
  Dropout Rate & 0.5 & \cite{c38}\\
    \hline
  \end{tabular}
   \label{tab11}
\end{table}

\subsubsection{ViT Hyperparameters:}
Vision Transformers (ViT) use specific hyperparameters to shape their architecture and tune performance in image classification tasks. These hyperparameters impact the model's learning process and its ability to generalize effectively to new data. Below are key hyperparameters that can be adjusted to enhance the performance of ViT models:

\begin{itemize}
    \item Number of layers: This parameter controls the depth of the model and its capacity to understand complex relationships within the image data. More layers can lead to a greater ability to capture intricate patterns.
    \item Embedding size: The size of the embedding space affects how the model represents input data. Larger embedding sizes provide more information but also increase model complexity.
    \item Patch size: In ViT, images are split into fixed-size patches. Patch size determines the granularity of the visual data, influencing the model's accuracy and learning process.
    \item Attention head size: The attention mechanism helps the model focus on relevant parts of the image. The head size influences how the model assigns attention across different areas, affecting performance.
    \item Dropout rate: Dropout is used to reduce overfitting by randomly dropping connections during training, thus encouraging better generalization.
    \item Activation function: Activation functions, such as ReLU or GELU, are used to process data in each layer. The choice of activation function can impact model performance.
    \item Optimizer: Optimizers like Adam, Adagrad, or RMSprop manage weight adjustments during training. The choice of optimizer and its associated parameters significantly affect model performance and efficiency.
    \item Number of attention heads: Controls how many parallel attention layers are used. Each head independently examines the input data. This hyperparameter plays a key role in how the model captures and interprets different features.
\end{itemize}

Fine-tuning hyperparameters is essential for achieving optimal performance and efficiency in ViT architectures. Adjusting these parameters can influence how effectively a model learns, its convergence rate, and its ability to generalize to new data. Customizing the model's behavior for specific tasks or datasets can lead to improved accuracy and robustness.

Table \ref{tab12} outlines the hyperparameters used in various ViT architectures with the Fashion MNIST dataset. This overview supports researchers and practitioners in making informed decisions when designing models, ensuring choices made enhance model performance and contribute to achieving superior outcomes.

\begin{table}[hbt!]
\centering
  \caption{Hyperparameters of ViT architectures in various studies}
\begin{tabular}{|p{3cm}|p{4cm}|p{5cm}|}
 \hline
 \textbf{Hyperparameters} & \textbf{Values} & \textbf{References} \\
 \hline
 Epochs  & 15, 10, 300, 30, 60& \cite{c45}, \cite{c46}, \cite{c47}, \cite{c49}, \cite{c51}\\
 \hline
 Embedding size & 768 and 102, 256 & \cite{c45}, \cite{c46} \\
   \hline
 Encoder block& 12 and 24, 6& \cite{c45}, \cite{c46} \cite{c47}\\
 \hline
 Number of attention heads &4 &\cite{c47} \\
 \hline
Batch size & 128, 256, 32& \cite{c46} \cite{c47},\cite{c49}, \cite{c51} \\
\hline
Weight decay& $3 * 10^2$, 0.03&\cite{c46}, \cite{c47} \\
 \hline
 Patch size & 64, 4 and 8, 4 & \cite{c45}, \cite{c46}, \cite{c47} \\
  \hline
 Activation function & GELU, Softmax & \cite{c45}, \cite{c45} \cite{c49}\\
 \hline
Learning rate  & $1e^4$, $5*10^4$, $5 * 10^4$, 0.1, 0.001&\cite{c45}, \cite{c47}, \cite{c49}, \cite{c51}\\
 \hline
 Dropout rate &0.1 &\cite{c47} \\
 \hline
 Optimizer & Adam & \cite{c45} \cite{c46} \cite{c49} \cite{c51}\\
 \hline
\end{tabular}
\label{tab12}
\end{table}

\subsection{Fashion Sector Particularities}
The incorporation of advanced image classification and object detection methods in the fashion industry has revolutionized the shopping experience for consumers while offering brands invaluable insights. Personalized recommendation systems leverage these technologies to provide individualized outfit suggestions based on a customer's tastes and preferences, enhancing their shopping experience and boosting sales by aligning recommendations with their style \cite{c74} \cite{c75}.
Virtual try-on applications enable customers to visualize how clothing and accessories will look on them in real-time, using object detection and classification to create an engaging and interactive online shopping experience \cite{c76} \cite{c77}\cite{c78}. This innovation increases customer confidence in purchases and can lead to lower return rates. Trend analysis tools utilize image recognition and classification to help fashion brands monitor and predict emerging trends by analyzing large datasets of fashion images. These insights support strategic decisions in product design and marketing by highlighting popular styles, colors, and patterns \cite{c79}.

Research in the field highlights the benefits of these applications, such as high-accuracy techniques for categorizing fashion items like clothes and accessories. This assists with inventory management and precise advertising. Object detection algorithms can also classify fashion items in images, aiding in trend forecasting and brand positioning.

The fashion industry has emerged as a pivotal area for image classification tasks within the realm of computer vision, outpacing other industries in terms of significance and complexity. This is partly due to the expanding size of datasets, with each new dataset featuring an increasing number of classes. These classes encompass a variety of subcategories, such as different sizes, genders, colors, brands, and other attributes. Furthermore, the dynamic nature of fashion trends means that classes within datasets evolve periodically.
These characteristics present both challenges and opportunities for improving image classification tasks. Classification models must adapt to the rapidly changing styles and trends within the fashion industry, which helps enhance their accuracy and efficiency over time. By staying current with the ever-shifting landscape of fashion, image classification systems can achieve greater success in this sector.
\subsection{Limitations and Challenges}
When comparing ViTs and CNNs for Fashion MNIST image classification, several limitations and challenges should be taken into account to evaluate the performance and suitability of each architecture.

Data scarcity can pose a challenge for training both ViTs and CNNs, as diverse and high-quality datasets are essential for building models that generalize effectively across various fashion items. Data augmentation or transfer learning may be needed when data is limited. Another concern is domain adaptation, as models trained on Fashion MNIST may not perform well on other fashion datasets due to differences in styles and trends. This is particularly relevant for ViTs, which rely on high-quality input data for effective attention-based processing. Model interpretability is a crucial consideration, especially given the opaque nature of deep learning models. Understanding how a model arrives at its classifications can build trust and offer insights into how each architecture processes fashion images, which is important in comparing ViTs and CNNs. Biases in data or models can also lead to incorrect classifications and the reinforcement of stereotypes. Minimizing biases and ensuring fair, accurate classifications are essential goals for both ViTs and CNNs. Managing computational complexity is another key challenge, particularly when balancing performance with resource usage. ViTs can be more computationally intensive than CNNs, which may limit their practicality for certain applications.

Despite these limitations and challenges, ongoing research and development in ViTs and CNNs for Fashion MNIST classification continue to advance the field, offering opportunities for more efficient and reliable models in the future.

\section{Conclusion}
After thoroughly examining the application of deep learning techniques, particularly CNNs and ViTs, in Fashion MNIST image classification tasks, it's evident that these methods have significantly propelled the field forward. The studies reviewed showcase the effectiveness of both CNNs and ViTs in accurately categorizing fashion items, achieving notable milestones in classification accuracy and model efficiency. However, despite their successes, several challenges persist, including computational complexity, sensitivity to hyperparameters, interpretability issues, and reliance on labeled data.

The discussion section delves into the strengths and weaknesses of CNNs and ViTs, emphasizing the necessity for ongoing research to tackle these limitations. Additionally, it explores the potential of hybrid approaches that combine CNNs and ViTs, suggesting promising avenues for future exploration. Furthermore, the discussion underscores the importance of enhancing model interpretability, robustness, and the utilization of unlabeled data to bolster model performance and versatility.

Looking ahead, it's crucial to concentrate on optimizing model architectures, improving interpretability, harnessing unlabeled data, fortifying models against adversarial attacks, and investigating novel hybrid approaches. These future trajectories hold significant potential for advancing Fashion MNIST classification and deep learning methodologies overall, facilitating the development of more effective, interpretable, and resilient models.

%
%

\begin{thebibliography}{99}
\bibitem{c1}
Cunha, Maria Nascimento, et al. "Digital Lens: Exploring Customer Perceptions in the World of E-Commerce and E-Marketplaces." (2024): 150-167.
\bibitem{c2}
Micu, Adrian, et al. "Assessing an on-site customer profiling and hyper-personalization system prototype based on a deep learning approach." Technological Forecasting and Social Change 174 (2022): 121289.
\bibitem{c3}
Heni, Ashref, Imen jdey, and Hela Ltifi. "Blood Cells Classification Using Deep Learning With Customized Data Augmentation and Using Deep Learning With Custmized Data Augmentation And EK-Means Segmentarion." Journal of Theoretical and Applied Information Technology 101.3 (2023): 1-12.
\bibitem{c4}
Deldjoo, Yashar, et al. "A review of modern fashion recommender systems." ACM Computing Surveys 56.4 (2023): 1-37.
\bibitem{c5}
Dargan, Shaveta, et al. "A survey of deep learning and its applications: a new paradigm to machine learning." Archives of Computational Methods in Engineering 27 (2020): 1071-1092.
\bibitem{c6}
Vijayaraj, A., et al. "Deep learning image classification for fashion design." Wireless Communications and Mobile Computing 2022 (2022).
\bibitem{c7}
Maurício, José, Inês Domingues, and Jorge Bernardino. "Comparing Vision Transformers and Convolutional Neural Networks for Image Classification: A Literature Review." Applied Sciences 13.9 (2023): 5521.
\bibitem{c8}
Lee, Chin Poo, et al. "Plant-CNN-ViT: plant classification with ensemble of convolutional neural networks and vision transformer." Plants 12.14 (2023): 2642.
\bibitem{c9}
Cong, Shuang, and Yang Zhou. "A review of convolutional neural network architectures and their optimizations." Artificial Intelligence Review 56.3 (2023): 1905-1969.
\bibitem{c10}
Slimani, Nawel, Imen Jdey, and Monji Kherallah. "Performance comparison of machine learning methods based on CNN for satellite imagery classification." 2023 9th International Conference on Control, Decision and Information Technologies (CoDIT). IEEE, 2023.
\bibitem{c11}
Hcini, G. H. A. Z. A. L. A., et al. "Hyperparameter optimization in customized convolutional neural network for blood cells classification." J. Theor. Appl. Inf. Technol 99 (2021): 5425-5435.
\bibitem{c12}
Krizhevsky, Alex, Ilya Sutskever, and Geoffrey E. Hinton. "ImageNet classification with deep convolutional neural networks." Communications of the ACM 60.6 (2017): 84-90.
\bibitem{c13}
Muhammad, Usman, et al. "Pre-trained VGGNet architecture for remote-sensing image scene classification." 2018 24th International Conference on Pattern Recognition (ICPR). IEEE, 2018.
\bibitem{c14}
Haripriya, P., G. Parthiban, and R. Porkodi. "A STUDY ON CNN ARCHITECTURE OF VGGNET AND RESNET FORDICOM IMAGE CLASSIFICATION." NeuroQuantology 20.16 (2022): 2027.
\bibitem{c15}
He, Kaiming, et al. "Deep residual learning for image recognition." Proceedings of the IEEE conference on computer vision and pattern recognition. 2016.
\bibitem{c16}
Qomariah, Dinial Utami Nurul, Handayani Tjandrasa, and Chastine Fatichah. "Classification of diabetic retinopathy and normal retinal images using CNN and SVM." 2019 12th International Conference on Information \& Communication Technology and System (ICTS). IEEE, 2019.
\bibitem{c17}
Iswanto, Irene Anindaputri, Amadeus Suryo Winoto, and Michael Kristianus. "Fruits Recognition using Deep Convolutional Neural Network for Low Computing Device." Engineering, MAthematics and Computer Science Journal (EMACS) 5.2 (2023): 85-91.
\bibitem{c18}
Jamil, Sonain, Md Jalil Piran, and Oh-Jin Kwon. "A comprehensive survey of transformers for computer vision." Drones 7.5 (2023): 287.
\bibitem{c19}
Dosovitskiy, Alexey, et al. "An image is worth 16x16 words: Transformers for image recognition at scale." arXiv preprint arXiv:2010.11929 (2020): 1-22.
\bibitem{c20}
Khan, Salman, et al. "Transformers in vision: A survey." ACM computing surveys (CSUR) 54.10s (2022): 1-41.
\bibitem{c21}
Bouzidi, Sonia, Imen Jdey, and Adel Alimi. "A Vision Transformer Approach with L2 Regularization for Sustainable Fashion Classification." Available at SSRN 4686032.
\bibitem{c22}
Aladhadh, Suliman, et al. "An effective skin cancer classification mechanism via medical vision transformer." Sensors 22.11 (2022): 4008.
\bibitem{c23}
Peng, Zhiliang, et al. "Conformer: Local features coupling global representations for visual recognition." Proceedings of the IEEE/CVF international conference on computer vision. 2021.
\bibitem{c24}
Dai, Zihang, et al. "Coatnet: Marrying convolution and attention for all data sizes." Advances in neural information processing systems 34 (2021): 3965-3977.
\bibitem{c25}
Yunusa, Haruna, et al. "Exploring the Synergies of Hybrid CNNs and ViTs Architectures for Computer Vision: A survey." arXiv preprint arXiv:2402.02941 (2024).
\bibitem{c26}
Xin, Jia, et al. "Convolutional Neural Network for Fashion Images Classification (Fashion-MNIST)." Journal of Applied Technology and Innovation (e-ISSN: 2600-7304) 7.4 (2023): 11.
\bibitem{c27}
An, Hyosun, et al. "Conceptual framework of hybrid style in fashion image datasets for machine learning." Fashion and Textiles 10.1 (2023): 1-18.
\bibitem{c28}
Wu, Hui, et al. "Fashion iq: A new dataset towards retrieving images by natural language feedback." Proceedings of the IEEE/CVF Conference on computer vision and pattern recognition. 2021.
\bibitem{c29}
Hu, Shell Xu, et al. "Pushing the limits of simple pipelines for few-shot learning: External data and fine-tuning make a difference." Proceedings of the IEEE/CVF Conference on Computer Vision and Pattern Recognition. 2022.
\bibitem{c30}
Treneska, Sandra, and Sonja Gievska. "Object detection and instance segmentation of fashion images." Conference for Informatics and Information Technology. 2020.
\bibitem{c31}
Wang, Xinhui. "Towards color compatibility in fashion using machine learning." (2019).
\bibitem{c32}
Bhatt, Dulari, et al. "CNN variants for computer vision: History, architecture, application, challenges and future scope." Electronics 10.20 (2021): 2470.
\bibitem{c33}
Rathore, Bharati. "Cloaked in Code: AI \& Machine Learning Advancements in Fashion Marketing." Eduzone: International Peer Reviewed/Refereed Multidisciplinary Journal 6.2 (2017): 25-31.
\bibitem{c34}
Xiao, Han, Kashif Rasul, and Roland Vollgraf. "Fashion-mnist: a novel image dataset for benchmarking machine learning algorithms." arXiv preprint arXiv:1708.07747 (2017).

\bibitem{c35}
LEITHARDT, VALDERI. "Classifying garments from fashion-MNIST dataset through CNNs." Advances in Science, Technology and Engineering Systems Journal 6.1 (2021): 989-994.
\bibitem{c36}
Khanday, Owais Mujtaba, Samad Dadvandipour, and Mohd Aaqib Lone. "Effect of filter sizes on image classification in CNN: A case study on CFIR10 and fashion-MNIST datasets." IAES International Journal of Artificial Intelligence 10.4 (2021): 872.
\bibitem{c37}
Nocentini, Olivia, et al. "Image classification using multiple convolutional neural networks on the fashion-MNIST dataset." Sensors 22.23 (2022): 9544.
\bibitem{c38}
Erkoç, Tuğba, and Mustafa Taner Eskıl. "A Novel Similarity Based Unsupervised Technique for Training Convolutional Filters." IEEE Access (2023).
\bibitem{c39}
Swain, Debabrata, et al. "An Intelligent Fashion Object Classification Using CNN." EAI Endorsed Transactions on Industrial Networks and Intelligent Systems 10.4 (2023).
\bibitem{c40}
Yu, Feng, et al. "FFENet: frequency-spatial feature enhancement network for clothing classification." PeerJ Computer Science 9 (2023): e1555.
\bibitem{c41}
Wan, Guangquan, and Lan Yao. "LMFRNet: A Lightweight Convolutional Neural Network Model for Image Analysis." Electronics 13.1 (2023): 129.
\bibitem{c42}
Sun, Yulin, et al. ”MADPL-net: Multi-layer attention dictionary pair learning network for image classification.” Journal of Visual Communication and Image Representation 90 (2023): 103728.
\bibitem{c43}
Shin, Seong-Yoon, Gwanghyun Jo, and Guangxing Wang. "A novel method for fashion clothing image classification based on deep learning." Journal of Information and Communication Technology 22.1 (2023): 127-148.
\bibitem{c44}
METLEK, Sedat, and Halit ÇETİNER. "CNNTuner: Image Classification with A Novel CNN Model Optimized Hyperparameters." Bitlis Eren Üniversitesi Fen Bilimleri Dergisi 12.3: 746-763.

\bibitem{c45}
Mukherjee, Prerana, Chandan Kumar Roy, and Swalpa Kumar Roy. "OCFormer: One-Class Transformer Network for Image Classification." arXiv preprint arXiv:2204.11449 (2022).
\bibitem{c46}
Chhabra, Sachin, et al. "PatchRot: A Self-Supervised Technique for Training Vision Transformers." arXiv preprint arXiv:2210.15722 (2022).
\bibitem{c47}
Chhabra, Sachin, Hemanth Venkateswara, and Baoxin Li. "PatchSwap: A Regularization Technique for Vision Transformers." (2022).

\bibitem{c48}
Bacochina, Giovanni Araujo, and Rodrigo Clemente Thom de Souza. ”Element-Wise Attention Layers: an option for optimization.” arXiv preprint arXiv:2302.05488 (2023): 1-26.
\bibitem{c49}
Abd Alaziz, Hadeer M., et al. "Enhancing Fashion Classification with Vision Transformer (ViT) and Developing Recommendation Fashion Systems Using DINOVA2." Electronics 12.20 (2023): 4263.
\bibitem{c50}
Rodriguez, David, and Ram Krishnan. "Learnable Image Transformations for Privacy Enhanced Deep Neural Networks." 2023 5th IEEE International Conference on Trust, Privacy and Security in Intelligent Systems and Applications (TPS-ISA). IEEE Computer Society, 2023.
\bibitem{c51}
Shah, S. Muhammad Ahmed Hassan, et al. "A Hybrid Neuro-Fuzzy Approach for Heterogeneous Patch Encoding in ViTs Using Contrastive Embeddings \& Deep Knowledge Dispersion." IEEE Access (2023).
\bibitem{c52}
Xu, Chen, et al. "Transformer in optronic neural networks for image classification." Optics \& Laser Technology 165 (2023): 109627.
\bibitem{c53}
Shao, Ran, and Xiao-Jun Bi. "Transformers meet small datasets." IEEE Access 10 (2022): 118454-118464.
\bibitem{c54}
Cools, Aurélie, Sidi Ahmed Mahmoudi, and Mohammed Amin Belarbi. "CARENET: A NOVEL ARCHITECTURE FOR LOW DATA REGIME MIXING CONVOLUTIONS AND ATTENTION." (2023).
\bibitem{c55}
Meng, Yanju, et al. "MixMobileNet: A Mixed Mobile Network for Edge Vision Applications." Electronics 13.3 (2024): 519.
\bibitem{c56}
Hassani, Ali, and Humphrey Shi. "Dilated neighborhood attention transformer." arXiv preprint arXiv:2209.15001 (2022).
\bibitem{c57}
Xu, Chenhao, et al. "HSViT: Horizontally Scalable Vision Transformer." arXiv preprint arXiv:2404.05196 (2024).
\bibitem{c58}
Statista: \url{https://www.wizishop.fr/blog/lancer-ecommerce.html}.
\bibitem{c59}
Data Report: \url{https://dash.app/blog/ecommerce-statistics}.
\bibitem{c60}
Bazi, Yakoub, et al. "Vision transformers for remote sensing image classification." Remote Sensing 13.3 (2021): 516.
\bibitem{c61}
Vilas, Martina G., Timothy Schaumlöffel, and Gemma Roig. "Analyzing Vision Transformers for Image Classification in Class Embedding Space." Advances in Neural Information Processing Systems 36 (2024).
\bibitem{c62}
Strudel, Robin, et al. "Segmenter: Transformer for semantic segmentation." Proceedings of the IEEE/CVF international conference on computer vision. 2021.
\bibitem{c63}
d’Ascoli, Stéphane, et al. "Convit: Improving vision transformers with soft convolutional inductive biases." International conference on machine learning. PMLR, 2021.
\bibitem{c64}
Zhou, Tianfei, et al. "Rethinking semantic segmentation: A prototype view." Proceedings of the IEEE/CVF Conference on Computer Vision and Pattern Recognition. 2022.
\bibitem{c65}
Linsley, Drew, et al. "Learning long-range spatial dependencies with horizontal gated recurrent units." Advances in neural information processing systems 31 (2018).
\bibitem{c66}
Yang, Jianwei, et al. "Focal attention for long-range interactions in vision transformers." Advances in Neural Information Processing Systems 34 (2021): 30008-30022.
\bibitem{c67}
Simonyan, Karen, and Andrew Zisserman. "Very deep convolutional networks for large-scale image recognition." arXiv preprint arXiv:1409.1556 (2014).
\bibitem{c68}
Zhong, Zilong, et al. "Squeeze-and-attention networks for semantic segmentation." Proceedings of the IEEE/CVF conference on computer vision and pattern recognition. 2020.
\bibitem{c69}
Dosovitskiy, Alexey, et al. "An image is worth 16x16 words: Transformers for image recognition at scale." arXiv preprint arXiv:2010.11929 (2020).
\bibitem{c70}
Deininger, Luca, et al. "A comparative study between vision transformers and cnns in digital pathology." arXiv preprint arXiv:2206.00389 (2022).
\bibitem{c71}
Bello, Irwan. "Lambdanetworks: Modeling long-range interactions without attention." arXiv preprint arXiv:2102.08602 (2021).
\bibitem{c72}
Zhai, Xiaohua, et al. "Scaling vision transformers." Proceedings of the IEEE/CVF conference on computer vision and pattern recognition. 2022.
\bibitem{c73}
Zalando is the Europe’s largest online fashion platform: \url{https://www.zalando.com/}
\bibitem{c74}
Chakraborty, Samit, et al. "Fashion recommendation systems, models and methods: A review." Informatics. Vol. 8. No. 3. MDPI, 2021.
\bibitem{c75}
Deldjoo, Yashar, et al. "A review of modern fashion recommender systems." ACM Computing Surveys 56.4 (2023): 1-37.
\bibitem{c76}
Yuan, Miaolong, et al. "A mixed reality virtual clothes try-on system." IEEE Transactions on Multimedia 15.8 (2013): 1958-1968.
\bibitem{c77}
ALZAMZAMI, Ohoud, et al. "Smart Fitting: An Augmented Reality mobile application for Virtual Try-On." Romanian Journal of Information Technology and Automatic Control 33.2 (2023): 103-118.
\bibitem{c78}
Kakade, Parul, et al. "Digital Apparel Try-Ons: A Technological Odyssey into the World of Virtual Dressing Rooms and Consumer Engagement." International Research Journal on Advanced Engineering and Management (IRJAEM) 2.02 (2024): 12-19.
\bibitem{c79}
Rudniy, Alex, Olena Rudna, and Arim Park. "Trend tracking tools for the fashion industry: the impact of social media." Journal of Fashion Marketing and Management: An International Journal (2023).
\bibitem{c80}
Jdey, Imen, Ghazala Hcini, and Hela Ltifi. "Deep learning and machine learning for Malaria detection: Overview, challenges and future directions." arXiv preprint arXiv:2209.13292 (2022).
\bibitem{c81}
Suo, Leiming, et al. "Wind speed prediction by a swarm intelligence based deep learning model via signal decomposition and parameter optimization using improved chimp optimization algorithm." Energy 276 (2023): 127526.
\bibitem{c82}
Ali, Yasser A., et al. "Hyperparameter search for machine learning algorithms for optimizing the computational complexity." Processes 11.2 (2023): 349.
\bibitem{c83}
Bischl, Bernd, et al. "Hyperparameter optimization: Foundations, algorithms, best practices, and open challenges." Wiley Interdisciplinary Reviews: Data Mining and Knowledge Discovery 13.2 (2023): e1484.
\bibitem{c84}
Rao, Jun, et al. "Watch and Buy: A Practical Solution for Real-time Fashion Product Identification in Live Stream." Proceedings of the 1st Workshop on Multimodal Product Identification in Livestreaming and WAB Challenge. 2021.
\bibitem{c85}
Pang, Shanchen, et al. "An efficient style virtual try on network for clothing business industry." arXiv preprint arXiv:2105.13183 (2021).
\bibitem{c86}
Meng, Xiaoxu, Weikai Chen, and Bo Yang. "Neat: Learning neural implicit surfaces with arbitrary topologies from multi-view images." Proceedings of the IEEE/CVF Conference on Computer Vision and Pattern Recognition. 2023.
\bibitem{c87}
Krichen, Moez. "Convolutional neural networks: A survey." Computers 12.8 (2023): 151.
\bibitem{c88}
Patel, Sanskruti. "A comprehensive analysis of Convolutional Neural Network models." International Journal of Advanced Science and Technology 29.4 (2020): 771-777.
\bibitem{c89}
Liu, Yang, et al. "A survey of visual transformers." IEEE Transactions on Neural Networks and Learning Systems (2023).
\bibitem{c90}
Yin, Hongxu, et al. "Adavit: Adaptive tokens for efficient vision transformer." arXiv preprint arXiv:2112.07658 (2021).
\bibitem{c91}
Touvron, Hugo, Matthieu Cord, and Hervé Jégou. "Deit iii: Revenge of the vit." European conference on computer vision. Cham: Springer Nature Switzerland, 2022.
\bibitem{c92}
Tang, Yehui, et al. "A Survey on Transformer Compression." arXiv preprint arXiv:2402.05964 (2024).
\bibitem{c93}
Yuan, Li, et al. "Tokens-to-token vit: Training vision transformers from scratch on imagenet." Proceedings of the IEEE/CVF international conference on computer vision. 2021.
\bibitem{c94}
Gong, Chengyue, and Dilin Wang. "NASViT: Neural architecture search for efficient vision transformers with gradient conflict-aware supernet training." ICLR Proceedings 2022 (2022).
\bibitem{c95}
Chen, Chun-Fu Richard, Quanfu Fan, and Rameswar Panda. "Crossvit: Cross-attention multi-scale vision transformer for image classification." Proceedings of the IEEE/CVF international conference on computer vision. 2021.
\bibitem{c96}
Touvron, Hugo, et al. "Going deeper with image transformers." Proceedings of the IEEE/CVF international conference on computer vision. 2021.
\bibitem{c97}
Caron, Mathilde, et al. "Emerging properties in self-supervised vision transformers." Proceedings of the IEEE/CVF international conference on computer vision. 2021.
\bibitem{C98}
Han, Kai, et al. "Transformer in transformer." Advances in neural information processing systems 34 (2021): 15908-15919.
\bibitem{C99}
Liu, Ze, et al. "Swin transformer: Hierarchical vision transformer using shifted windows." Proceedings of the IEEE/CVF international conference on computer vision. 2021.
\bibitem{c100}
Rajaraman, Sivaramakrishnan, et al. "Pre-trained convolutional neural networks as feature extractors toward improved malaria parasite detection in thin blood smear images." PeerJ 6 (2018): e4568.
\bibitem{c101}
Strisciuglio, Nicola, Manuel Lopez-Antequera, and Nicolai Petkov. "Enhanced robustness of convolutional networks with a push–pull inhibition layer." Neural Computing and Applications 32 (2020): 17957-17971.
\bibitem{c102}
Keele, Staffs. "Guidelines for performing systematic literature reviews in software engineering." (2007).
\bibitem{C103}
Brahmi, Walid, and Imen Jdey. "Automatic tooth instance segmentation and identification from panoramic X-Ray images using deep CNN." Multimedia Tools and Applications (2023): 1-21.
\bibitem{C104}
\url{https://keras.io/api/applications/}
\bibitem{c105}
Liu, Zhenhua, et al. "Post-training quantization for vision transformer." Advances in Neural Information Processing Systems 34 (2021): 28092-28103.
\end{thebibliography}
%

\end{document}